\documentclass[pdflatex,sn-mathphys-num]{sn-jnl}

\usepackage{graphicx}
\usepackage{multirow}
\usepackage{amsmath,amssymb,amsfonts}
\usepackage{amsthm}
\usepackage{mathrsfs}
\usepackage[title]{appendix}
\usepackage{xcolor}
\usepackage{textcomp}
\usepackage{manyfoot}
\usepackage{booktabs}
\usepackage{array}
\usepackage{algorithm}
\usepackage{algorithmic}
\usepackage{listings}

\begin{document}

\title[Article Title]{From Pixels to Prompts: Physics-Guided Language Enhancement for AI-generated Image Detection}


\author{\fnm{Mei} \sur{Qiu}}\email{qiumei@300188.cn}

% \author[1]{\fnm{Jianqiang} \sur{Zhao}}\email{zhaojq@300188.cn}


\affil{\orgdiv{Artificial Intelligence Research Institute}, \orgname{SDIC Intelligence Co., Ltd.}, \orgaddress{\street{Software Park Phase III}, \city{Xiamen}, \postcode{361024}, \state{Fujian}, \country{China}}}

% \affil[2]{\orgdiv{School of Informatics}, \orgname{Xiamen University}, \orgaddress{\street{422 Siming South Road}, \city{Xiamen}, \postcode{361005}, \state{Fujian}, \country{China}}}

% \affil[3]{\orgdiv{Department}, \orgname{Organization}, \orgaddress{\street{Street}, \city{City}, \postcode{610101}, \state{State}, \country{Country}}}


\abstract{AI-generated content (AIGC) has substantially narrowed the visual gap between natural and synthetic images, while existing detectors often rely on generator-specific artifacts and consequently exhibit limited generalization to unseen generative models. This motivates a fundamental question: which intrinsic physical image characteristics provide stable and discriminative cues for distinguishing natural from AI-generated images? In this work, we investigate this question from two perspectives: systematically identifying robust physical features and incorporating them into a vision--language representation without relying on external textual prompts at inference. We first conduct a comprehensive analysis of 15 handcrafted physical descriptors across images generated by more than 20 generative models. Based on their cross-generator stability and real--fake discriminability, we propose a Feature Stability--Discriminability Validation Assessment (FSDVA) for selecting a compact set of reliable physical features. Five core features, including Laplacian variance, Sobel statistics, and residual-noise variance, are subsequently used to construct structured textual descriptions that encode physical evidence for CLIP-based learning. We integrate these physics-enriched descriptions with image representations through a native physics-guided multimodal framework, adapting CLIP ViT-L/14 with LoRA. During inference, the model requires only the input image and does not depend on external captions or manually provided prompts. On the GenImage benchmark, our method achieves 99.0\% accuracy on SDv1.4 and 96.7\% on Wukong, with a mean accuracy of 97.23\% across eight generative models, compared with 95.8\% for C2pCLIP under the same evaluation setting. On UniversalFakeDetect, which evaluates generalization across diverse and unseen generation settings, the proposed method maintains strong and stable detection performance. These results demonstrate that carefully selected physical image statistics can provide complementary forensic evidence to pretrained vision--language representations, offering an interpretable and more generator-agnostic approach to AI-generated image detection.

}




\maketitle

\section{Introduction}\label{sec1}
Recent advances in generative artificial intelligence have substantially improved the visual quality and diversity of synthetic images. Early generative adversarial networks (GANs) demonstrated increasingly realistic image synthesis \cite{goodfellow2014generative,karras2019style,karras2020analyzing}, while diffusion models, such as DDPM \cite{ho2020denoising} and latent diffusion models \cite{rombach2022high}, further improved photorealism and controllability. Large-scale generative models have subsequently narrowed the perceptual gap between real and synthetic imagery \cite{ramesh2022hierarchical,team2023gemini}. This rapid proliferation of synthetic content has raised concerns regarding media authenticity and digital misinformation \cite{saetra2023generative,zhang2025unmasking}, motivating reliable AI-generated image detection.

Existing AIGC image detectors can be broadly categorized into data-driven visual detectors and representation-driven or feature-driven approaches \cite{xu2025recent,zhang2025unmasking}. Data-driven methods learn discriminative representations directly from real and generated images, often exploiting frequency or other image-level artifacts \cite{wang2020cnnfingerprint,frank2020leveraging,dzanic2020fourier,wang2023dire}. However, detectors trained on specific generators may degrade substantially on unseen generative models \cite{ojha2023towards,zhu2023genimage,wang2023dire}. Large-scale studies further show that increasing the diversity of generators during training can improve generalization, highlighting the importance of cross-generator evaluation \cite{park2025community}.

Another line of research exploits pretrained vision-language representations. CLIP \cite{radford2021learning} provides transferable visual features with strong generalization to unseen generative models \cite{ojha2023towards}. Subsequent approaches further adapt CLIP through prompt learning and multimodal alignment \cite{tan2025c2p}. Nevertheless, semantic representations are primarily optimized for high-level visual concepts and may not explicitly capture subtle image-formation artifacts. This raises an important question: \emph{can low-level forensic evidence complement foundation-model representations to improve cross-generator generalization?}

Complementary forensic studies have shown that generated images exhibit systematic discrepancies in frequency spectra, residual signals, and other low-level statistics \cite{dzanic2020fourier,frank2020leveraging,jeong2022frepg}. These artifacts arise partly because real photographs pass through a physical imaging pipeline involving optics, sensors, demosaicing, and image signal processing, whereas generative models optimize image distributions without explicitly reproducing this complete acquisition process. Such discrepancies therefore provide complementary evidence for distinguishing generated from camera-captured images \cite{dzanic2020fourier,jeong2022frepg}.

However, existing physical-feature-based methods remain fragmented, with different studies focusing on individual cues such as frequency statistics, residuals, or noise patterns \cite{dzanic2020fourier,frank2020leveraging,jeong2022frepg}. Moreover, many descriptors are evaluated on limited generator sets, making it difficult to determine whether they capture generator-independent forensic properties. Recent benchmarks such as GenImage explicitly introduce cross-generator evaluation \cite{zhu2023genimage}, while Community Forensics demonstrates that both the scale and diversity of generator training data are important for detector generalization \cite{park2025community}. These observations motivate a systematic investigation of physical descriptors under heterogeneous generative mechanisms.

Motivated by this gap, we investigate two questions: \textbf{(Q1) Which physical descriptors provide stable evidence across different generators?} and \textbf{(Q2) Can such physical evidence complement semantic representations to improve cross-generator detection?}

To investigate the physical cues that distinguish real and AI-generated images, we first compile 15 representative physical descriptors from the forensic literature, covering frequency, noise, residual, texture, color, and related image statistics. We systematically analyze these descriptors across more than 20 generative models under a cross-generator evaluation protocol, and use the resulting stability and discriminability measurements to identify a compact set of reliable physical features. Based on these findings, we develop a native physics-guided multimodal framework that incorporates the selected physical evidence into pretrained CLIP through structured visual-conditioned representations, enabling low-level forensic cues to complement high-level visual semantics for cross-generator AIGC detection.

Our main contributions are summarized as follows:
\begin{itemize}
\item \textbf{Comprehensive analysis of physical features.} We conduct a systematic empirical study of 15 representative physical descriptors sourced from the forensic literature across more than 20 generative models, providing a cross-generator analysis of their stability and discriminative characteristics.

\item \textbf{Physical-feature selection algorithm.} We develop a feature stability--discriminability based selection algorithm that jointly evaluates the cross-generator stability and real--fake discriminability of physical descriptors, enabling the identification of a compact and reliable subset of forensic features.

\item \textbf{Native physics-guided detection framework.} We propose a native physics-guided multimodal framework that incorporates the selected physical evidence into pretrained CLIP through structured visual-conditioned representations, integrating low-level forensic cues with high-level visual semantics to improve cross-generator AIGC detection.
\end{itemize}

\section{Related Works}
We review existing literature on synthetic detection (including face forgery and general AIGC detection) by categorizing methodologies into two principal domains: data-driven and physics-based approaches.

\textbf{Data-driven methods} dominate the field, leveraging deep learning to distinguish real from synthetic content. Early work by Rossler et al.~\cite{rossler2019faceforensics++} employed Xception for face manipulation detection, while Wang et al.~\cite{wang2021representative} addressed limited model focus via attention-based augmentation. To improve generalization, Cao et al.~\cite{cao2022end} proposed a reconstruction-classification framework, Chen et al.~\cite{chen2022self} utilized augmented forgeries with adversarial training, and Shiohara and Yamasaki~\cite{shiohara2022detecting} introduced SBIs for robust representations. With the advent of foundation models, Ojha et al.~\cite{ojha2023towards} leveraged pre-trained vision-language features in a non-learning approach, Liu et al.~\cite{liu2024forgery} developed FatFormer to integrate frequency features, and Tan et al.~\cite{tan2025c2p} proposed C2P-CLIP for detection. Recent advances include disentanglement frameworks~\cite{yan2023ucf}, latent space augmentation~\cite{yan2024transcending}, and fusion of co-occurrence matrices with CNNs~\cite{nataraj2019detecting}. This evolution reflects the field's progression from CNN-based artifact mining to modern pre-trained and disentanglement approaches.

\textbf{Physics-based and physics-informed methods} leverage inherent physical constraints—such as frequency spectra, spatial texture, gradients, and temporal dynamics—to identify stable synthetic traces. Frank et al.~\cite{frank2020leveraging} showed GAN upsampling introduces consistent frequency artifacts, while Chai et al.~\cite{chai2020makes} found local patch artifacts persist even in refined generators. Liu et al.~\cite{liu2020global} used global texture statistics via Gram-Net, and Haliassos et al.~\cite{haliassos2021lips} proposed LipForensics targeting lip movement dynamics. Subsequent work introduced re-synthesis frameworks~\cite{he2021beyond}, frequency filtering~\cite{jeong2022bihpf}, adaptive texture-frequency mining~\cite{liu2023adaptive}, and gradient-based representations~\cite{tan2023learning}. Tan et al.~\cite{tan2024rethinking} improved generalization by modeling upsampling inconsistencies, while Cheng et al.~\cite{cheng2025co} fused pixel artifacts with semantic anomalies. For video detection, Zhang et al.~\cite{zhang2025physics} proposed a normalized spatiotemporal gradient (NSG) quantifying deviations from natural dynamics. Unlike data-driven methods, physics-based approaches offer stronger robustness and generalization by relying on fundamental constraints rather than dataset-specific patterns, though they often require domain expertise to design effective features.
\begin{figure}[tb]
  \centering
  \includegraphics[height=8.5cm]{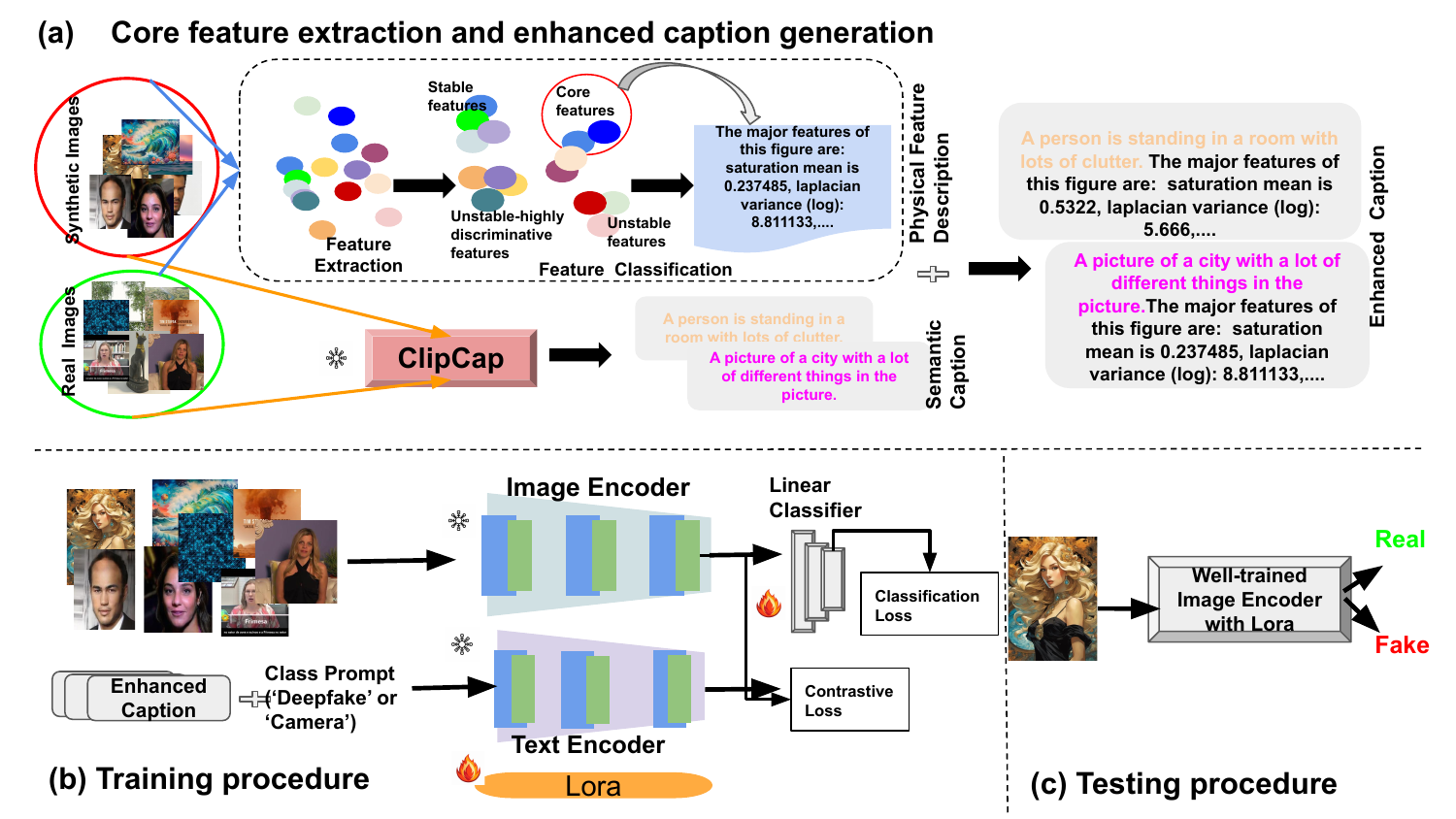}
  \caption{\small Overall workflow of the proposed synthetic image detection framework. (a) Enhanced caption preparation: Integrate input images’ core physical features into captions to enrich representation. (b) Training: Use enhanced captions and class prompts to train the model for real/AI-generated image discrimination. (c) Testing: Input images into the trained model to predict AI generation. \textbf{ Our training pipeline framework natively supports end-to-end training without additional architectural adjustments.}}
  \label{fig:whole_frame}
\end{figure}

\section{Methods}
Inspired by FakeBench \cite{li2025fakebench}, Explainable MLLM \cite{li2025token}, and C2P-CLIP \cite{tan2025c2p}, we propose and comprehensively evaluate 15 classic physical features. These 15 discriminative visual attributes are computed to quantify statistical and perceptual discrepancies between authentic and AI-generated imagery. Each feature targets distinct artifacts inherent in synthetic content—such as unnatural smoothness, color inconsistency, or repetitive patterns—with mathematical formulations tailored to amplify these differences and normalize feature ranges for robust model input. We design a new algorithm to estimate the stability and discriminative power of each feature. We then select core features that exhibit both high stability and strong discrimination and incorporate these features into text descriptions. The enhanced captions are formed by merging the original ClipCap \cite{mokady2021clipcap} generated texts with these new core feature-based descriptions. We adopt the training and testing pipeline from C2P-CLIP to investigate whether core features-based descriptions can improve feature understanding and benefit fake image detection. The overall framework is illustrated in Fig. \ref{fig:whole_frame}.

\subsection{Feature Definition and Extraction}
Each feature is normalized or log-scaled to stabilize its dynamic range. The 15 features are formally defined as in Table \ref{tab:features_min_space}: % Add to preamble
% No extra packages needed (ECCV template has amsmath pre-included)
% Wrap the align environment in a font-size command + spacing tweaks
% No extra packages needed (ECCV template supports tabular/amsmath natively)
\begin{table*}[tb]
\centering
\fontsize{7}{8}\selectfont % Exact font size: 7pt text, 8pt line height (adjust 7 to 6/8 for fine-tuning)
\setlength{\tabcolsep}{0pt} % Zero column padding
\renewcommand{\arraystretch}{0.15} % Minimal vertical spacing
\setlength{\mathsurround}{0pt} % No math padding
\begin{tabular}{@{}ccc@{}} % No left/right table padding
\toprule
$F_1\!=\!\frac{1}{HW}\sum_{i,j}\text{HSV}_{i,j,1}$ &
$F_2\!=\!\frac{1}{HW}\sum_{i,j}\text{HSV}_{i,j,2}$ &
$F_3\!=\!\log\!\left(1+\operatorname{Var}(\nabla^2 I_g)\right)$ \\
\midrule
$F_4\!=\!10\!\cdot\!\frac{1}{HW}\sum_{i,j}\sqrt{S_x^2\!+\!S_y^2}$ &
$F_5\!=\!10\!\cdot\!\sqrt{\frac{1}{HW}\sum_{i,j}(M_{i,j}\!-\!\mu_M)^2}$ &
$F_6\!=\!\log\!\left(1+\frac{1}{N}\sum_k\operatorname{Var}(\operatorname{DCT}(B_k))\right)$ \\
\midrule
$F_7\!=\!\log\!\left(1+1000\operatorname{Var}(R)\right)$ &
$F_8\!=\!-\sum_b p_b\log_2 p_b$ &
$F_9\!=\!\frac{\operatorname{Cov}(R,G)}{\sqrt{\operatorname{Var}(R)\operatorname{Var}(G)}}$ \\
\midrule
$F_{10}\!=\!\frac{\operatorname{Cov}(R,B)}{\sqrt{\operatorname{Var}(R)\operatorname{Var}(B)}}$ &
$F_{11}\!=\!-\sum_v p_v\log_2 p_v$ &
$F_{12}\!=\!-\sum_v p_v\log_2 p_v$ \\
\midrule
$F_{13}\!=\!\log\!\left(1+\operatorname{Var}(\text{HSV}_{:,:,0})\right)$ &
$F_{14}\!=\!\frac{1}{HW}\sum_{i,j}\frac{E_{i,j}}{255}$ &
$F_{15}\!=\!\log\!\left(1+|\operatorname{Kurt}(B)|\right)$ \\
\bottomrule
\end{tabular}
\vspace{-1pt}
\caption{\tiny Mathematical definitions of 15 synthetic image detection features. Key symbol definitions: $i,j$ = 2D pixel coordinates (height/width); $H,W$ = image height/width; $\text{HSV}_{i,j,c}$ = HSV color space value at pixel $(i,j)$ for channel $c$ (1=saturation, 2=brightness, 0=hue); $I_g$ = grayscale image; $\nabla^2$ = Laplacian operator (second-order derivative for sharpness); $S_x/S_y$ = Sobel horizontal/vertical edge responses; $M_{i,j}$ = Sobel edge magnitude at $(i,j)$; $\mu_M$ = mean edge magnitude; $\operatorname{DCT}(\cdot)$ = Discrete Cosine Transform; $B_k$ = $k$-th 8×8 image block; $N=1000$ (sampled blocks for efficiency); $R$ = residual noise (image minus non-local means denoised version); $p_b$ = normalized frequency of Local Binary Pattern (LBP) bin $b$; $\operatorname{Cov}(\cdot,\cdot)$ = covariance; $\operatorname{Var}(\cdot)$ = variance; $R/G/B$ = red/green/blue color channels; $p_v$ = normalized pixel value frequency for chrominance channels (Cr/Cb); $E_{i,j}$ = Canny edge detector output (0=non-edge, 255=edge); $\operatorname{Kurt}(\cdot)$ = kurtosis (tailedness of intensity distribution).}
\label{tab:features_min_space}
\end{table*}
\begin{align}
\mathrm{CV}(f)
&=
\frac{\sigma_{\mu_f}}{\mu_{\mu_f}},
\label{eq:cv}\\
S_s(f)
&=
\max\left(0,\min\left(1,1-\mathrm{CV}(f)\right)\right),
\label{eq:stability}\\
\mathrm{JMD}_k(f)
&=
\frac{
\left|\mu_{\mathrm{real}}^k(f)-\mu_{\mathrm{fake}}^k(f)\right|
}{
\sigma_{\mathrm{real}}^k(f)+\sigma_{\mathrm{fake}}^k(f)
},
\label{eq:jmd}\\
\widetilde{\mathrm{JMD}}(f)
&=
\frac{\overline{\mathrm{JMD}}(f)}
{1+\overline{\mathrm{JMD}}(f)},
\label{eq:jmd_norm}\\
S_d(f)
&=
\frac{
\widetilde{\mathrm{JMD}}(f)
+
\overline{\mathrm{AUC}}(f)
}{2}.
\label{eq:discriminability}
\end{align}

\small
where $\sigma_{\mu_f}$ and $\mu_{\mu_f}$ denote the standard deviation and mean, respectively, of the dataset-wise means of feature $f$; $\mu_{\mathrm{real}}^k(f)$ and $\mu_{\mathrm{fake}}^k(f)$ denote the means of feature $f$ for real and fake samples in dataset $\mathcal{D}*k$, respectively; and $\sigma*{\mathrm{real}}^k(f)$ and $\sigma_{\mathrm{fake}}^k(f)$ denote their corresponding standard deviations. $\overline{\mathrm{JMD}}(f)$ and $\overline{\mathrm{AUC}}(f)$ denote the mean JMD and AUC of feature $f$ across datasets, respectively, while $\widetilde{\mathrm{JMD}}(f)$ is the normalized JMD mapped to $[0,1)$.

\begin{algorithm}[tb]
\small
\caption{FSDVA (Feature Stability-Discriminability Validation Assessment)}
\label{alg:fsdva}
\begin{algorithmic}[1]
\REQUIRE $D$: Dictionary of dataset-wise features
$\{\mathcal{D}_k:
\{\mathbf{F}_{\mathrm{real}}^k \in \mathbb{R}^{N_k\times15},
\mathbf{F}_{\mathrm{fake}}^k \in \mathbb{R}^{N_k\times15}\}\}$
\ENSURE $\mathcal{M}$: Metric table containing $S_s(f)$, $S_d(f)$,
and the class assigned to each feature $f$

\FOR{each feature $f \in \{f_1,\ldots,f_{15}\}$}

    \STATE \textbf{Step 1: Compute Stability Score} $S_s(f)$
    \STATE For each dataset $\mathcal{D}_k$, compute the dataset-wise mean
    $\mu_f^k$ of $f$ using the combined real and fake samples
    \STATE Compute the cross-dataset coefficient of variation
    $\mathrm{CV}(f)$ using Eq.~\ref{eq:cv}
    \STATE Compute
    $S_s(f)=\max\left(0,\min\left(1,1-\mathrm{CV}(f)\right)\right)$
    using Eq.~\ref{eq:stability}

    \STATE \textbf{Step 2: Compute Discriminability Score} $S_d(f)$
    \FOR{each dataset $\mathcal{D}_k$}
        \STATE Compute the inter-/intra-class distance ratio
        $\mathrm{JMD}_k(f)$ using Eq.~\ref{eq:jmd}
        \STATE Train a Logistic Regression classifier using only $f$
        and compute $\mathrm{AUC}_k(f)$
    \ENDFOR
    \STATE Compute mean JMD and AUC across datasets:
    $\overline{\mathrm{JMD}}(f)$ and $\overline{\mathrm{AUC}}(f)$
    \STATE Normalize the mean JMD:
    $\widetilde{\mathrm{JMD}}(f)=
    \overline{\mathrm{JMD}}(f)/
    \left(1+\overline{\mathrm{JMD}}(f)\right)$
    using Eq.~\ref{eq:jmd_norm}
    \STATE Compute
    $S_d(f)=
    \left(\widetilde{\mathrm{JMD}}(f)
    +\overline{\mathrm{AUC}}(f)\right)/2$
    using Eq.~\ref{eq:discriminability}

    \STATE \textbf{Step 3: Assign Feature Class}
    \STATE
    $\mathrm{Class}(f)=
    \begin{cases}
    \mathrm{Core\ Feature},
    & S_s(f)\geq0.7 \land S_d(f)\geq0.5,\\
    \mathrm{Usable\ Feature},
    & S_s(f)\geq0.45 \land S_d(f)\geq0.3,\\
    \mathrm{Unstable\ High\mbox{-}Discrim},
    & S_s(f)<0.7 \land S_d(f)\geq0.4,\\
    \mathrm{Unusable\ Feature},
    & \mathrm{otherwise}.
    \end{cases}$

    \STATE Append
    $\{f,S_s(f),S_d(f),\mathrm{Class}(f)\}$ to $\mathcal{M}$
\ENDFOR
\RETURN $\mathcal{M}$
\end{algorithmic}
\label{algo:FSDVA}
\end{algorithm}

The 15 handcrafted features are:
\emph{saturation mean} ($F_1$), \emph{brightness mean} ($F_2$),
\emph{Laplacian variance} ($F_3$), \emph{Sobel magnitude mean} ($F_4$),
\emph{Sobel magnitude standard deviation} ($F_5$),
\emph{DCT variance} ($F_6$), \emph{residual noise variance} ($F_7$),
\emph{LBP entropy} ($F_8$), \emph{RG correlation} ($F_9$),
\emph{RB correlation} ($F_{10}$), \emph{chroma entropy dim-1} ($F_{11}$),
\emph{chroma entropy dim-2} ($F_{12}$), \emph{hue variance} ($F_{13}$),
\emph{Canny edge density} ($F_{14}$), and \emph{blue-channel kurtosis} ($F_{15}$).

\subsection{Feature Stability and Discriminability Evaluation}

To systematically assess the 15 handcrafted features for synthetic image detection, we propose the Feature Stability-Discriminability Validation Assessment (FSDVA) in Algorithm~\ref{alg:fsdva}. FSDVA evaluates each feature from two complementary perspectives: \emph{cross-dataset stability} and \emph{real--fake discriminability}.

For stability, FSDVA first computes the mean of each feature over the combined real and fake samples within each dataset, and then calculates the coefficient of variation (CV) across the resulting dataset-wise means using Eq.~\ref{eq:cv}. The CV is converted into a normalized stability score $S_s(f)$ using Eq.~\ref{eq:stability}, where a higher score indicates more consistent feature statistics across different datasets.

For discriminability, FSDVA evaluates each feature independently on every dataset using two complementary measures: the inter-/intra-class distance ratio $\mathrm{JMD}_k(f)$ defined in Eq.~\ref{eq:jmd}, and the AUC of a Logistic Regression classifier trained using only that feature to distinguish real from fake images. The JMD and AUC values are averaged across datasets to obtain $\overline{\mathrm{JMD}}(f)$ and $\overline{\mathrm{AUC}}(f)$, respectively. Since JMD is unbounded whereas AUC lies in $[0,1]$, the mean JMD is first mapped to $[0,1)$ using Eq.~\ref{eq:jmd_norm}. The normalized JMD and mean AUC are then averaged to obtain the discriminability score $S_d(f)$ in Eq.~\ref{eq:discriminability}.

Finally, each feature is assigned to one of four categories according to its $(S_s,S_d)$ scores: \emph{Core Feature}, \emph{Usable Feature}, \emph{Unstable High-Discrim}, or \emph{Unusable Feature}. The thresholds were determined prior to the final evaluation based on a combination of validation-set experiments, the observed distributions in the feature analysis, findings from related literature, and empirical experience with handcrafted forensic features. Specifically, $S_s(f)\geq0.7$ and $S_d(f)\geq0.5$ are adopted as relatively stringent criteria for identifying Core Features, requiring both strong cross-dataset stability and moderate-to-strong discriminability. Lower thresholds of $S_s(f)\geq0.45$ and $S_d(f)\geq0.3$ are used to retain features with weaker but still practically useful properties as Usable Features. Features satisfying $S_s(f)<0.7$ but $S_d(f)\geq0.4$ are separately categorized as Unstable High-Discrim features, reflecting features that exhibit relatively strong discriminative capability but insufficient cross-dataset stability. Importantly, these thresholds were determined using validation and exploratory feature-analysis results and were not optimized using the final test set.

\begin{figure}[tb]
  \centering
\includegraphics[width=1.0\textwidth]{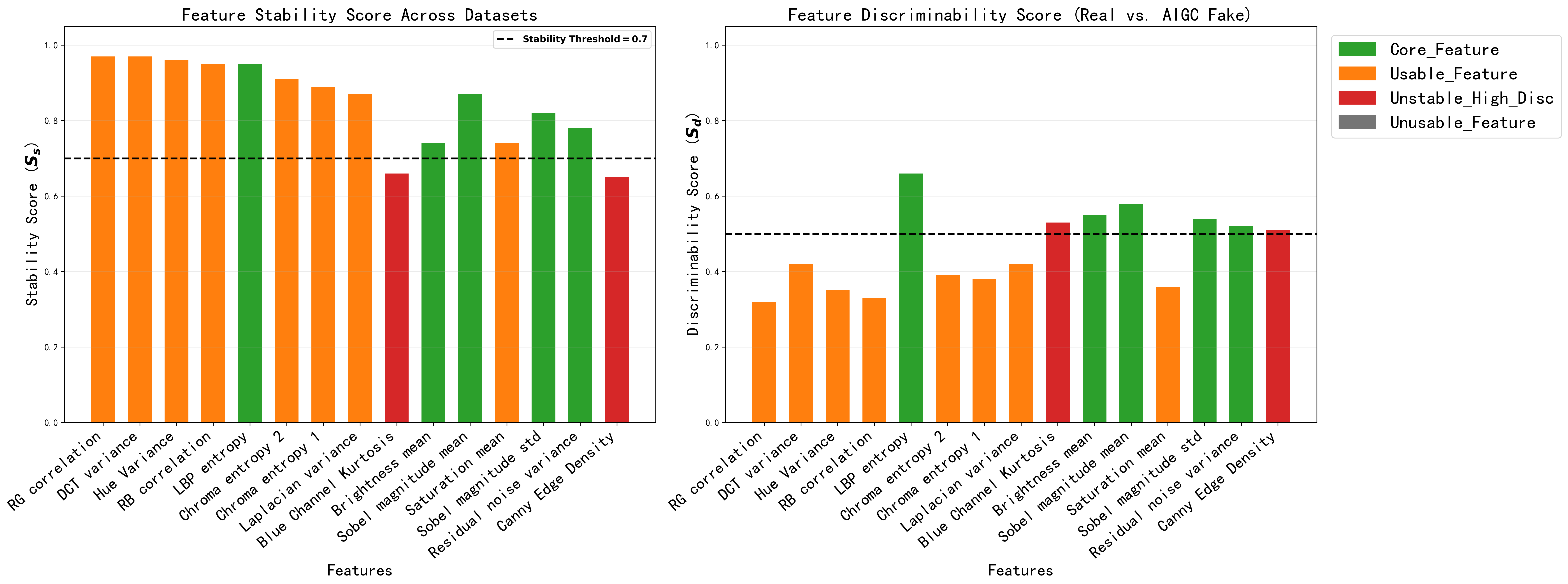}
  \caption{\small Stability ($S_s$, left) and discriminability ($S_d$, right) scores of image features for fake image detection. Based on the thresholds, set features to Core Feature(green), Usable Feature(orange), red Unstable High-Discrim(red), and Unusable Feature(grey).}
% \label{fig:feature_metrics}.
  \label{fig:feat_cls}
\end{figure}
\begin{figure}[tb]
  \centering \includegraphics[width=1.0\textwidth]{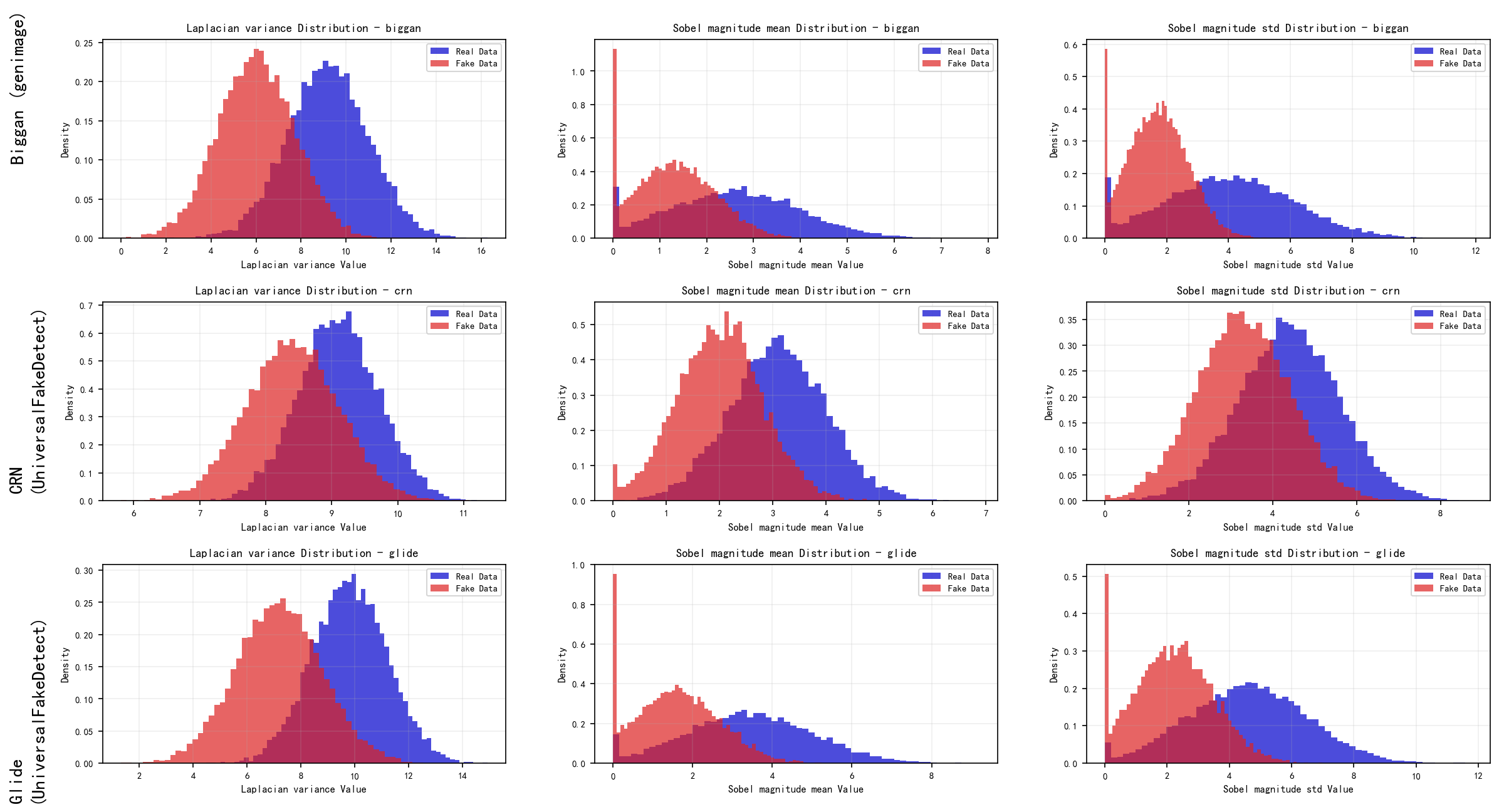}
  \caption{\small Density distributions of core features from original images (Laplacian variance, Sobel magnitude mean/std over BigGAN/CRN/Glide (UniversalFakeDetect). Blue = real images, red = fake images.}
  \label{fig:core_feat}
\end{figure}
\begin{figure}[tb]
  \centering
\includegraphics[width=0.95\textwidth]{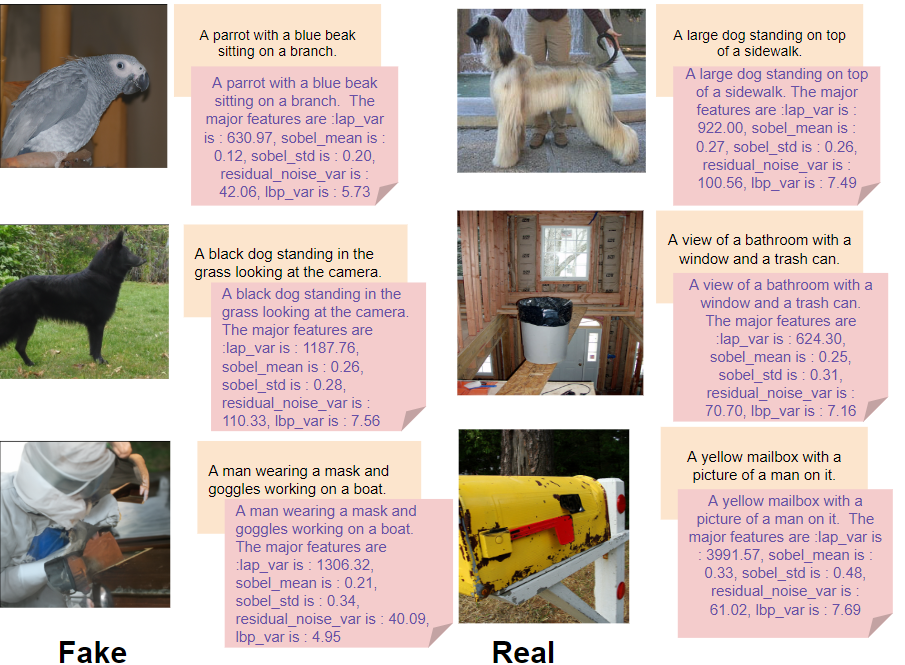}
  \caption{\small Image-text pairs with original image captions and enhanced by core features' description.}
% \label{fig:feature_metrics}.
  \label{fig:cap_samples}
\end{figure}

\begin{table*}[t]
\centering
\resizebox{\linewidth}{!}{
\begin{tabular}{lcccccccccc}
\toprule
Method & Ref & Midjourney & SDv1.4 & SDv1.5 & ADM & GLIDE & Wukong & VQDM & BigGAN & mAcc \\
\midrule
UniFD & CVPR 2023 \cite{ojha2023towards} & 93.9 & 96.4 & 96.2 & 71.9 & 85.4 & 94.3 & 81.6 & 90.5 & 88.8 \\
NPR & CVPR 2024 \cite{tan2024rethinking} & 81.0 & 98.2 & 97.9 & 76.9 & 89.8 & 96.9 & 84.1 & 84.2 & 88.6 \\
FreqNet & AAAI 2024 \cite{tan2024frequency} & 89.6 & 98.8 & 98.6 & 66.8 & 86.5 & 87.3 & 75.8 & 81.4 & 86.8 \\
FatFormer & CVPR 2024 \cite{liu2024forgery} & 92.7 & \textbf{100.0} & \textbf{99.9} & 75.9 & 88.0 & \textbf{99.9} & \textbf{98.8} & 55.8 & 88.9 \\
C2pClip (Trump.Biden) & AAAI 2025 \cite{tan2025c2p} & 82.2 & 95.1 & 95.5 & \textbf{95.1} & 98.9 & 98.7 & 93.8 & 98.3 & 94.7 \\
C2pClip (Deepfake.Camera) & AAAI 2025 \cite{tan2025c2p} & 88.2 & 90.9 & 97.9 & 96.4 & \textbf{99.8} & 98.8 & 96.5 & \textbf{98.7} & 95.8 \\
IAPL & CVPR 2026 \cite{li2026towards} & 95.1 & 99.9 & 99.8 & 85.4 & 95.9 & 99.8 & 98.8 & 98.7 & 96.7\\
\midrule
\textbf{Ours (Deepfake.Camera)} & -- & \textbf{97.4} & 99.0 & 96.7 & 96.4 & 97.0 & 96.7 & 96.1 & 98.3 & \textbf{97.23} \\
\bottomrule
\end{tabular}
}
\caption{Comparison with state-of-the-art methods on the GenImage dataset, where the best confidence threshold for each method is selected on an isolated 20\% validation split by maximizing the F1-score and then fixed when reporting test accuracy. The highest accuracy in each column is highlighted in bold. mAcc denotes the average accuracy across all evaluated generative models.}
\label{tab:genimage_cross_model_acc}
\end{table*}
\begin{table*}[t]
\centering
\resizebox{\linewidth}{!}{
\begin{tabular}{lcccccccccc}
\toprule
\multirow{2}{*}{Method} & \multirow{2}{*}{Ref} & DALLE& \multicolumn{3}{c}{Glide} & ADM & \multicolumn{3}{c}{LDM} & \multirow{2}{*}{mAcc} \\
\cmidrule(lr){3-5} \cmidrule(lr){6-8} \cmidrule(lr){9-11}
& & 100\_10 & 100\_27 & 50\_27 & 100\_10 & 100\_27 & 50\_27 & 100 & 200 & 200\_cfg \\
\midrule
CNNDet\cite{wang2020cnn} & CVPR 2020 & 51.8  & 53.3 & 53.0 & 54.2 & 54.9 & 51.9 & 52.0 & 51.6 & 52.8 \\
FreDect \cite{frank2020leveraging} & ICML 2020 & 57.0 & 53.6 & 50.4 & 52.0 & 53.4 & 56.6 & 56.4 & 56.5 & 54.5 \\
F3Net \cite{qian2020thinking} & ECCV 2020 & 71.6& 88.3 & 87.0 & 88.5 & 69.2 & 74.1 & 73.4 & 80.7 & 79.1 \\
LGrad [28] & CVPR 2023 & 88.5 & 89.4 & 87.4 & 90.7 & 86.6 & 94.8 & 94.2 & 95.9 & 90.9 \\
UnivFD  \cite{ojha2023towards} & CVPR 2023 & 89.5& 90.1 & 90.7 & 91.1 & 75.7 & 90.5 & 90.2 & 77.3 & 86.9 \\
PatchCraft \cite{zhong2311patchcraft} & Arxiv 2023 & 83.3 &80.1 & 83.4 & 77.6 & 80.9 & 88.9 & 89.3 & 88.1 & 84.0 \\
FreqNet\cite{tan2024frequency}  & AAAI 2024 & 97.2& 87.8 & 84.4 & 86.6 & 67.2 & 97.8 & 97.4 & 97.2 & 89.5 \\
NPR  \cite{tan2024rethinking} & CVPR 2024 & 94.5& 98.2 & 97.8 & 98.2 & 75.8 & \textbf{99.3} & 99.1 & \textbf{99.0} & 95.1 \\
FatFormer \cite{liu2024forgery} & CVPR 2024 & \textbf{98.7} & 94.5 & 94.1 & 94.2 & 75.9 & 98.5 & 98.5 & 94.8 & 93.6 \\
SAFE\cite{li2025improving} & KDD 2025 & 97.5 & 97.2 & 95.8 & 96.6 & 82.4 & 98.8 & 98.8 & 98.7 & 95.7 \\
COD \cite{jia2025secret} & CVPR 2025 & 98.1 & 97.6 & 97.4 & 97.7 & 92.8 & 98.8 & \textbf{98.9} & 98.7 & 97.5 \\

PPM-CLIP \cite{wang2026ppm} & CVPR 2026 & 98.4& 99.3& 99.3 & 99.3 & 97.0 & \textbf{99.3}& 99.1 & 99.1 & 98.8 \\
\midrule
Ours & -- & \textbf{99.25} &  \textbf{99.3} & \textbf{99.35} & \textbf{99.35} & \textbf{9.97} & 99.35 & \textbf{99.35} & \textbf{99.35} & \textbf{99.41} \\
\bottomrule
\end{tabular}
}
\caption{\textbf{Comparison with state-of-the-art methods on the test subsets of the UniversalFakeDetect benchmark in terms of Accuracy (\%).} The evaluated settings cover four generation sources, including DALLE, Glide, ADM, and LDM, with different sampling configurations as specified in the column headers. The best result in each setting is highlighted in bold. PPM-CLIP~\cite{wang2026ppm}, a recent CVPR 2026 detection method, is included as a contemporary baseline. The mean accuracy (mAcc) is computed across all evaluated settings.}

\label{tab:ojha_sota_comparison}
\end{table*}
\subsection{Train Clip with Enhanced Caption by Merging Core Features}
As shown in Fig.~\ref{fig:whole_frame}, our method follows the same training and testing pipeline as C2pClip~\cite{tan2025c2p}, with the only modification being the enrichment of input captions with core physical features extracted via FSDVA (Algorithm.~\ref{algo:FSDVA}). 

Given a training set \( X = \{x_j, y_j\}_{j=1}^N \) with \( y \in \{0,1\} \) indicating fake (1) or real (0), each image \( x_j \) has an associated caption \( c_j \) generated by ClipCap. We compute a core feature set \( \mathcal{F}(x_j) \) and convert it into a descriptive text \( f_j \). The enhanced caption is formed by concatenation: \( \widetilde{c}_j = c_j \oplus f_j \). 

Text and image features are extracted via \( u_j = \text{encoder}_{\text{text}}(\widetilde{c}_j) \) and \( v_j = \text{encoder}^{\text{Lora}}_{\text{img}}(x_j) \). The training objective combines contrastive and classification losses:
\begin{equation}
   \mathcal{L} = \mathcal{L}_{\text{contrastive}} + \alpha \cdot \mathcal{L}_{\text{classification}}, 
\end{equation}
where \( \mathcal{L}_{\text{contrastive}} \) is the symmetric cross-entropy loss between paired image-text embeddings. At test time, only the image encoder with LoRA and the classifier are used for detection:
\begin{equation}
   p = \text{classifier}\left( \text{encoder}^{\text{Lora}}_{\text{img}}(x) \right). 
\end{equation}

\section{Experiments}
In this section, we provide the evaluation including datasets, implementation details, and deepfake synthetic performance.
\subsection{Experiments Settings}
\vspace{-2mm}
\smallskip
\noindent
\textbf{Data.} Our feature-extraction data are randomly sampled from the test subsets of GenImage~\cite{zhu2023genimage} and UniversalFakeDetect~\cite{ojha2023towards}. GenImage contains eight synthetic-image subsets generated by Midjourney, SDv1.4, SDv1.5, ADM, GLIDE, Wukong, VQDM, and BigGAN. From UniversalFakeDetect, we select ten representative subsets, including BigGAN, StarGAN, GauGAN, Deepfake, CRN, IMLE, Guided Diffusion, LDM, GLIDE, and DALL-E. Thus, our feature analysis covers 18 dataset subsets in total, corresponding to 16 unique generation/manipulation sources after accounting for the overlap of BigGAN and GLIDE between the two datasets. For the synthetic-image detection experiments, both the training and testing data are drawn exclusively from GenImage.

\smallskip
\noindent
\textbf{Implementation Details.}
In the feature analysis stage, we randomly sample 6,000 real and 6,000 fake images from each GenImage subset, and 1,000-2,000 real--fake pairs from each selected UniversalFakeDetect subset. These samples are used exclusively for post-hoc analysis of the discriminative properties of the predefined features and are not used for feature selection, model training, validation, hyperparameter tuning, or final test evaluation. For each feature, we train a Logistic Regression classifier and report the corresponding AUC, with the regularization coefficient fixed to $C=1$ using the default solver. For feature-based descriptions, the extracted features are converted into textual descriptions using the template defined in Fig.~\ref{fig:cap_samples}, with numerical values rounded to the specified precision. These descriptions are generated offline before model training and remain fixed throughout all detection experiments.

For synthetic-image detection, we adopt the CLIP ViT-L/14 backbone with LoRA following C2pClip~\cite{tan2025c2p}. LoRA is applied to the specified attention layers with rank $r=32$ and scaling factor $\alpha_{\mathrm{LoRA}}=0.6$. We replace the original image captions with our feature-based descriptions while keeping the remaining CLIP architecture unchanged. The hyperparameter $\alpha$ in Eq.~(5) is set to $0.5$. The model is optimized using Adam with a learning rate of $2\times10^{-4}$ and a batch size of 128 for one epoch on four NVIDIA RTX 5090 GPUs. All input images are resized to $224\times224$. The confidence threshold is determined on the validation set and fixed before evaluation on the test set.

\smallskip
\noindent
\textbf{Performance Evaluation Metric.}
For feature selection, we evaluate candidate features by analyzing their cross-dataset distributions, domain robustness, and classification performance (AUC and JMD values) to identify core discriminative features. In the testing phase, we use mean accuracy (mAcc) as the primary evaluation metrics to quantify synthetic detection performance.

% \begin{table}[t]
% \centering
% \begin{minipage}{0.35\linewidth}
% \centering
% \begin{tabular}{lc}
% \toprule
% Method & Acc (\%) \\
% \midrule
% CLIP only & 92.3 \\
% CLIP + random features & 92.1 \\
% CLIP + all 15 features & 95.7 \\
% \textbf{CLIP + FSDVA selected features (Ours)} & \textbf{97.2} \\
% \bottomrule
% \end{tabular}
% \caption{Ablation study of physical feature integration on the GenImage dataset.}
% \label{tab:ablation_feature_integration}
% \end{minipage}
% \hfill
% \begin{minipage}{0.35\linewidth}
% \centering
% \begin{tabular}{lc}
% \toprule
% Number of Selected Features & Acc (\%) \\
% \midrule
% 1 feature (top-1) & 93.2 \\
% 3 features (top-3) & 94.8 \\
% 5 features (top-5) & \textbf{97.2} \\
% 10 features (top-10) & 96.2 \\
% 15 features (all) & 95.7 \\
% \bottomrule
% \end{tabular}
% \caption{Ablation study on the number of FSDVA-selected physical descriptors on the GenImage dataset. The best performance is achieved with 5 carefully selected features, indicating that redundant or unstable descriptors degrade detection accuracy.}
% \label{tab:ablation_feature_number}
% \end{minipage}
% \end{table}
\begin{table}[!htb]
    \begin{minipage}{.5\linewidth}
      \centering
        \begin{tabular}{ll}
          \toprule
Method & Acc (\%) \\
\midrule
CLIP only & 92.3 \\
CLIP + random features & 92.1 \\
CLIP + all 15 features & 95.7 \\
\textbf{CLIP + FSDVA selected features (Ours)} & \textbf{97.2} \\
\bottomrule
        \end{tabular}
        \caption{Ablation study of physical feature integration on the GenImage dataset.}
\label{tab:ablation_feature_integration}
    \end{minipage}%
    \begin{minipage}{.5\linewidth}
      \centering
        \begin{tabular}{ll}
         \toprule
Number of Selected Features & Acc (\%) \\
\midrule
1 feature (top‑1) & 93.2 \\
3 features (top‑3) & 94.8 \\
5 features (top‑5) & \textbf{97.2} \\
10 features (top‑10) & 96.2 \\
15 features (all) & 95.7 \\
\bottomrule
        \end{tabular}
      \caption{Ablation study on the number of FSDVA‑selected physical descriptors on the GenImage dataset. The best performance is achieved with 5 carefully selected features, indicating that redundant or unstable descriptors degrade detection accuracy.}
\label{tab:ablation_feature_number}
    \end{minipage} 
\end{table}

\subsection{Results}
% \smallskip
% \noindent
% \textbf{Features cross datasets.}

\smallskip
\noindent
\textbf{Feature Classification and Core Features.}
We extract 15 physical features from balanced real–fake image pairs and evaluate them using the Feature Stability and Discriminability Validity Analysis (FSDVA) framework. Based on stability (\(S_s\)) and discriminability (\(S_d\)) scores, we select 5 core features: LBP entropy, Laplacian variance, Sobel magnitude mean, Sobel magnitude std and residual noise variance that satisfy both high stability and strong discriminability thresholds.

As shown in Fig.~\ref{fig:feat_cls}, the stability and discriminability scores of all 15 candidate features are visualized, with color-coded categories: green for core features (high \(S_s\) and \(S_d\)), orange for usable features, red for unstable but highly discriminative features, and grey for unusable features. This selection process ensures that the core features are robust to dataset variations while effectively distinguishing between real and fake images. Fig.~\ref{fig:core_feat} presents the density distributions of these three core features across three datasets (BigGAN, CRN, and Glide) and it shows its strong stability cross generators and datasets. We observe that the observed patterns are robust to dataset variations, validating the utility of these core features for downstream fake image detection.

 \smallskip
\noindent
\textbf{Robustness and Inference Cost.}
To further assess practical robustness, we evaluate the proposed detector under common image transformations, including JPEG compression and image resizing at different degradation levels. The same transformations are applied to the test images while keeping the trained model and decision threshold fixed, and performance is compared with that on the original images. We additionally report the relative performance retention under each degradation condition to quantify robustness against common post-processing operations.

For computational efficiency, we measure inference time under a standardized hardware and software environment using the same input resolution and batch size for all compared methods. Before timing, several warm-up iterations are performed, and the reported inference time is averaged over repeated runs to reduce measurement noise. We report the average inference time per image together with the number of trainable parameters and model memory consumption, providing a direct comparison between detection performance and computational cost.

\smallskip
\noindent
\textbf{Ablation Studies} 
Ablation results in Tables 5 and 6 verify our FSDVA feature‑selection pipeline. Table \ref{tab:ablation_feature_integration} shows that direct integration of random or full‑set physical features cannot match the performance of CLIP combined with FSDVA‑filtered features (97.2\%). Table \ref{tab:ablation_feature_number} reveals that detection accuracy peaks using only five selected descriptors; performance drops when more features are incorporated, since redundant unstable artifacts introduce noise. This validates that FSDVA effectively picks a compact, high‑quality feature subset for AIGC detection.
\section{Discussion}
\smallskip
\noindent
\textbf{Why Physical Features Become Language Tokens?}
Physical features, such as Laplacian variance, Sobel statistics, and residual noise variance, provide valuable authenticity cues but are difficult to directly integrate into vision-language models due to the representation gap between low-level physical statistics and semantic features. We therefore convert these physical measurements into structured textual descriptions and use them as language prompts for CLIP. Rather than assuming that CLIP explicitly understands the numerical meaning of each physical measurement, we view the textual descriptions as a structured representation that encodes heterogeneous physical cues within the pretrained language embedding space. This formulation provides a unified interface for incorporating low-level forensic evidence into a vision-language model, allowing physical statistics to complement image-level semantic representations. The observed performance improvements suggest that such physics-enriched descriptions provide useful complementary information for distinguishing real and generated images, although the precise mechanism through which CLIP utilizes these numerical physical cues remains an interesting direction for further investigation.

\section{Conclusion}
We propose a physics-guided multimodal framework that incorporates physically grounded image statistics into language-based representations for AI-generated image detection. By introducing a feature selection strategy to identify stable and discriminative physical attributes, our method converts these attributes into structured textual descriptions and integrates them into the pretrained CLIP embedding space. Experimental results demonstrate that physics-enriched descriptions provide complementary forensic information and improve detection performance and generalization across the evaluated generation settings.

Despite these promising results, several limitations remain. First, it remains unclear whether CLIP explicitly captures the numerical semantics of individual physical measurements or primarily benefits from the structured information conveyed by the descriptions. Second, the fixed token length of the CLIP text encoder limits the number and granularity of physical attributes that can be incorporated, while representing each attribute with a single scalar may overlook localized and fine-grained forensic patterns. Third, the interaction between class prompts and physical-feature descriptions has not been fully explored. Finally, our evaluation is limited to the current benchmark datasets and a finite set of generative models, leaving the robustness of the framework under more diverse and unseen synthesis methods and real-world image transformations to be further validated.

Future work will investigate the relationship between numerical physical measurements and their language representations through controlled embedding-space and interpretability analyses. We will also explore more flexible multimodal architectures, including direct physical-token integration and cross-modal attention, as well as localized physical descriptors for capturing spatially varying generation artifacts. In addition, broader evaluations on unseen generative models and real-world image manipulations will be conducted to further improve the robustness, generalization, and interpretability of physics-guided AI-generated image detection.

\backmatter
\bibliography{sn-bibliography}

\end{document}